\documentclass[letterpaper]{article} % DO NOT CHANGE THIS
\usepackage{aaai2026}  % DO NOT CHANGE THIS
\usepackage{times}  % DO NOT CHANGE THIS
\usepackage{helvet}  % DO NOT CHANGE THIS
\usepackage{courier}  % DO NOT CHANGE THIS
\usepackage[hyphens]{url}  % DO NOT CHANGE THIS
\usepackage{graphicx} % DO NOT CHANGE THIS
\usepackage{natbib}  % DO NOT CHANGE THIS AND DO NOT ADD ANY OPTIONS TO IT
\usepackage{caption} % DO NOT CHANGE THIS AND DO NOT ADD ANY OPTIONS TO IT
\usepackage{algorithm}
\usepackage{algorithmic}
\usepackage{amsthm,amsmath,amssymb} % ljh 1
\newcommand{\equalmark}{*}
\newcommand{\leadmark}{\ensuremath{\dagger}}
\newcommand{\corrmark}{\ensuremath{\ddagger}}

\usepackage{xcolor}
\definecolor{printred}{cmyk}{0,1,1,0}

\usepackage{multirow} % for table
\usepackage{booktabs}
\usepackage{newfloat}
\usepackage{listings}
\DeclareCaptionStyle{ruled}{labelfont=normalfont,labelsep=colon,strut=off} % DO NOT CHANGE THIS
\floatstyle{ruled}
\newfloat{listing}{tb}{lst}{}
\floatname{listing}{Listing}
\title{VGOcc: Learning Visual-Geometric Gaussians for Vision-Centric\\
3D Driving Occupancy Prediction}

\author{
    Junhong Lin\textsuperscript{1,\equalmark},
    Xianda Guo\textsuperscript{2,\equalmark,\leadmark},
    Kangli Wang\textsuperscript{1},
    Yuqi Ye\textsuperscript{1},
    Xiaoyu Liang\textsuperscript{1},
    Yanlun Peng\textsuperscript{3},
    Wei Gao\textsuperscript{1,\corrmark}
}

\affiliations{
    \textsuperscript{1}SECE, Peking University\\
    \textsuperscript{2}School of Computer Science, Wuhan University\\
    \textsuperscript{3}Great Wall Motor\\
    jhlin42in@stu.pku.edu.cn,
    xianda\_guo@163.com,
    gaowei262@pku.edu.cn
}
\usepackage{bibentry}
\begin{document}
\nocopyright
\maketitle

\begingroup
\renewcommand{\thefootnote}{\fnsymbol{footnote}}
\footnotetext[1]{Equal contribution. 
\textsuperscript{\ensuremath{\dagger}}Project leader. 
\textsuperscript{\ensuremath{\ddagger}}Corresponding author.}
\endgroup

% ===========================================================================
% ==== Abstract ====
% ===========================================================================

\begin{abstract}

Vision-only occupancy prediction requires recovering a semantic 3D occupancy field from calibrated surround-view images, where each view provides observations with ambiguous depth along camera rays.
Existing methods have progressed from dense structured representations to sparse Gaussian primitives, improving the efficiency of 3D scene representation. 
However, Gaussian learning still relies primarily on image domain features, which provide limited explicit geometric information for volumetric reasoning.
Our key observation is that effective Gaussian occupancy modeling requires not only sparse primitives, but also richer geometric and semantic learning cues.
In this paper, we propose VGOcc, which learns visual and geometric cues from foundation models for Gaussian modeling. 
VGOcc incorporates these cues into primitive initialization and refinement, yielding a representation termed \emph{Visual-Geometric Gaussians} tailored to semantic occupancy prediction.
Specifically, we propose Visual-Geometric Gaussian Birth to form spatially balanced Gaussian centers from ray depth hypotheses, while visual semantic features initialize primitive attributes.
Next, we design Pose-Aware Feature Learning to combine foundation tokens with camera embeddings and calibrated ray information. Features from neighboring views are then aggregated at projected 3D locations for each Gaussian refinement stage.
Finally, Gaussian decoder refines birth Gaussians with pose-aware features and renders them into semantic occupancy.
Experiments on nuScenes demonstrate that VGOcc achieves state-of-the-art performance in vision-only 3D occupancy prediction.
Codes will be available at \url{https://github.com/JHLin42in/VGOcc}.

\end{abstract}

% ===========================================================================
% ==== 1 Introduction ====
% ===========================================================================

\section{Introduction}

Vision-centric 3D occupancy prediction provides a unified scene representation for autonomous driving. Given calibrated surround-view images, it estimates a dense semantic field in the ego coordinate frame, covering diverse objects, road surfaces, thin structures, and background regions. Such representations also support occupancy forecasting, cooperative perception, scene prediction with world models, and motion planning \citep{wang2025uniocc,yang2025driveoccworld,li2025preworld, zhu2026bridging3dgaussianssemantic,zhu2026gem}. 
However, cameras observe only visible surfaces rather than the complete 3D scene structure. Each image location defines a ray but does not reveal the occupied depth or structures hidden behind visible surfaces. Therefore, the key challenge of vision-only occupancy prediction is to infer a complete 3D semantic field from 2D ray observations with ambiguous depth.

% ==== Figure 1 ====
\begin{figure}[t]
  \centering
  \includegraphics[width=0.47\textwidth]{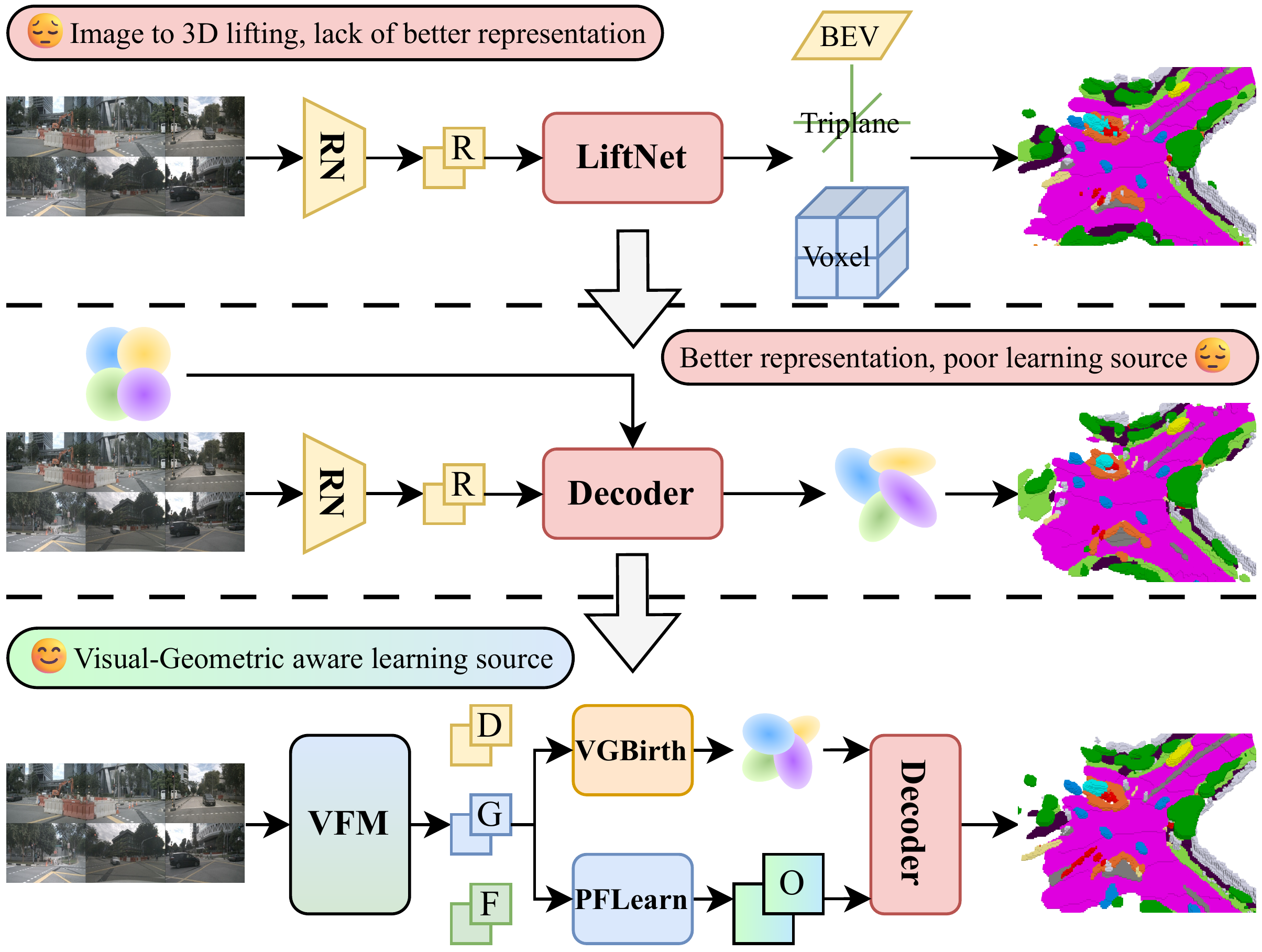}
  \caption{From structured lifting to \emph{Visual-Geometric Gaussians}.
  BEV, tri-plane, and voxel methods establish structured 3D representations, while sparse Gaussians reduce spatial redundancy.
  VGOcc learns \emph{Visual-Geometric Gaussians} by grounding their initialization and refinement in complementary visual and geometric cues.}
  \label{fig:teaser}
\end{figure}

Existing vision-centric occupancy methods mainly address this problem by lifting image features into structured 3D spaces. 
BEV based methods such as FB-OCC \citep{li2023fbocc} aggregate multiview features in bird's eye view, but height compression limits fine vertical modeling. 
Tri-plane methods extend BEV features across multiple orthogonal planes to preserve more complete spatial structure \citep{huang2023tpvformer}, while voxel methods build 3D feature volumes to capture fine geometry \citep{wei2023surroundocc,zhang2023occformer,li2023voxformer,ma2024cotr}.
These methods provide explicit 3D feature spaces but still allocate considerable computation to empty or weakly observed regions. 
GaussianFormer \citep{huang2024gaussianformer} and GaussianFormer-2 \citep{huang2025gaussianformer2} instead encode scenes with sparse semantic Gaussians, substantially reducing spatial redundancy and improving the efficiency of 3D scene representation.

Nevertheless, sparse Gaussian primitives alone do not resolve the limited explicit geometry available in image domain features, as illustrated in Figure~\ref{fig:teaser}.
Gaussian primitives are typically initialized using learned priors or image features, and subsequently refined by sampling image features at projected 3D reference points.
Although this provides visual information for Gaussian refinement, sampling at projected locations does not explicitly model the geometric relations required for volumetric reasoning.
This limitation constrains volumetric structure recovery and semantic modeling, particularly in occluded and distant regions.

We argue that improving occupancy prediction requires not only an efficient 3D Gaussian representation, but also richer learning of geometry and semantics.
In particular, visual and geometric foundation models provide rich semantic and geometric information \citep{oquab2024dinov2,wang2025vggt}. 
However, since they are pretrained for image and surface understanding, their priors cannot be directly exploited by Gaussian primitives for semantic occupancy reasoning.
Therefore, we aim to transform their informations into a useful learning cues for Gaussian primitives, yielding a representation we term \emph{Visual-Geometric Gaussians}.

To this end, we propose VGOcc, a sparse Gaussian occupancy framework that learns visual and geometric cues from foundation models and incorporates them into primitive initialization and refinement. 
First, Visual-Geometric Gaussian Birth (VGBirth) constructs the initial Gaussians from current scene. It predicts ray depth hypotheses from frozen geometric features and applies fast voxel-thinned sampling to obtain spatially balanced Gaussian centers. Visual semantic features along the corresponding rays initialize the primitive attributes, while learned fallback queries cover regions with weak geometric observations. 
Second, Pose-Aware Feature Learning (PFLearn) constructs features for iterative Gaussian refinement. It conditions and fuses DINO, frame, and global tokens with camera embeddings and calibrated ray information. The resulting features are aggregated across neighboring views at projected 3D locations for each refinement stage.
Finally, Gaussian decoder refines birth Gaussians with these features and renders the resulting \emph{Visual-Geometric Gaussians} into semantic occupancy. In this way, VGOcc preserves the efficiency of sparse Gaussian modeling while grounding both Gaussian birth and refinement in scene specific visual and geometric cues. Our main contributions are summarized as follows:

\begin{itemize}

\item We introduce VGOcc, which learns complementary visual and geometric cues from foundation models for Gaussian initialization and refinement, yielding a representation termed \emph{Visual-Geometric Gaussians} tailored to semantic occupancy prediction.

\item We propose Visual-Geometric Gaussian Birth, which forms spatially balanced Gaussian centers from ray depth hypotheses and initializes primitive attributes with visual semantic features.

\item We design Pose-Aware Feature Learning, which conditions and fuses foundation tokens using camera embeddings and calibrated ray information, then aggregates neighboring view features at projected 3D locations for iterative Gaussian refinement.

\item Extensive experiments on nuScenes demonstrate that VGOcc achieves state-of-the-art performance among the compared vision-only occupancy methods.

\end{itemize}

% ===========================================================================
% ==== 2 Related Work ====
% ===========================================================================

\section{Related Work}

\subsection{Vision-Centric Occupancy Prediction}

Vision-centric occupancy prediction has evolved from BEV perception to dense scene completion. BEVFormer aggregates multi-camera features with BEV transformers \citep{li2022bevformer}. TPVFormer introduces orthogonal planes to preserve vertical structure \citep{huang2023tpvformer}. SurroundOcc constructs dense multi-scale 3D volumes \citep{wei2023surroundocc}, while OccFormer adopts dual-path voxel transformers \citep{zhang2023occformer}. VoxFormer generates depth-guided sparse proposals before densification \citep{li2023voxformer}. Efficient variants such as COTR and SparseOcc reduce volumetric computation through compressed tokens or sparse latent modeling \citep{ma2024cotr,liu2024sparseocc}. Occ3D and OpenOccupancy establish semantic occupancy benchmarks for autonomous driving \citep{tian2023occ3d,wang2023openoccupancy}, while Humanoid-OmniOcc extends full-view occupancy evaluation to embodied AI \citep{guo2026humanoid}.

Sparse Gaussian methods replace dense voxel features with continuous primitives following 3D Gaussian Splatting \citep{kerbl2023gaussians}. GaussianFormer and GaussianFormer-2 adapt semantic 3D Gaussians to vision-centric occupancy prediction \citep{huang2024gaussianformer,huang2025gaussianformer2}. VG3T improves multi-view Gaussian generation through structured sampling and positional refinement \citep{kim2025vg3t}, while VG3S adapts frozen foundation features for Gaussian occupancy learning \citep{yan2026vg3s}. Unlike these methods, we learn complementary visual and geometric cues for Gaussian initialization and refinement.

\subsection{Visual and Geometric Foundation Models}

Visual and geometric foundation models provide transferable information for downstream 3D perception. DINOv2 learns robust appearance and semantic features through large-scale visual pretraining \citep{oquab2024dinov2}. DUSt3R predicts dense point maps directly from image pairs \citep{wang2024dust3r}, while MASt3R improves geometric matching and reconstruction with dense local descriptors \citep{leroy2024mast3r}. VGGT further unifies camera, depth, point map, and point track prediction within a feed-forward multi-view framework \citep{wang2025vggt}.

% ==== Figure 2 ====
\begin{figure*}[!t]
  \centering
	\includegraphics[width=1\textwidth]{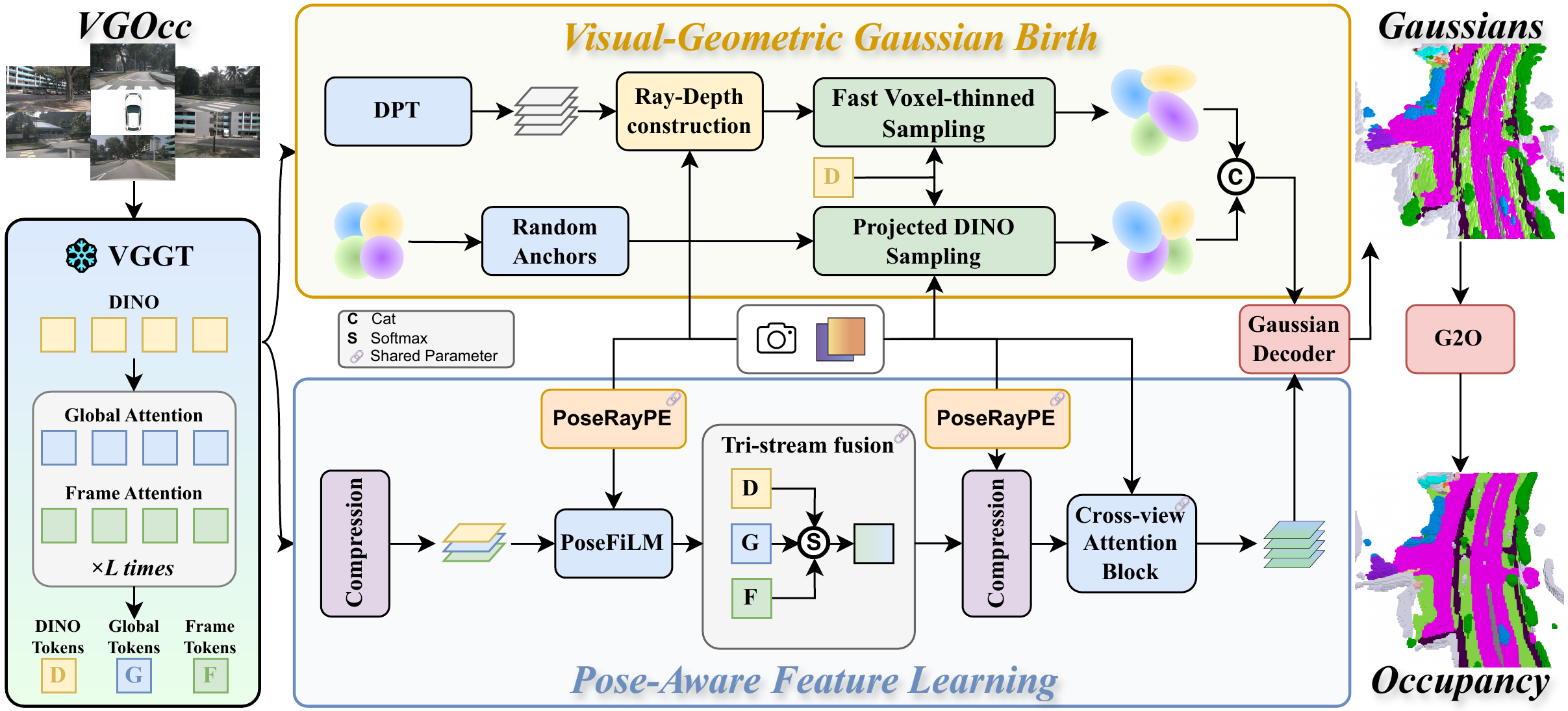}
  \caption{Overall architecture of VGOcc. Frozen VGGT provides complementary visual and geometric tokens. Visual-Geometric Gaussian Birth constructs Gaussian queries with geometry based initialization and visual semantic attribute priors, while Pose-Aware Feature Learning converts foundation tokens into pose-conditioned multi-view features. The Gaussian decoder refines the queries and renders the final semantic occupancy field.}
  \label{fig:method}
\end{figure*}

Recent work adapts these models to driving tasks. DriveVGGT, DVGT, DVGT-2 and DGGT extend geometry prediction to driving \citep{jia2025drivevggt,zuo2025dvgt,zuo2026dvgt2,chen2025dggt}. VGD applies geometry priors to surround-view Gaussian reconstruction \citep{lin2025vgd}. SURDS evaluates spatial reasoning in driving \citep{guo2025surds}. AutoDrive-P3 unifies perception, prediction and planning \citep{ye2026autodrivep3}. OmniNWM studies navigation world models \citep{li2026omninwm}. SparseOccVLA and VGGDrive connect occupancy or geometry with language-guided planning \citep{dang2026sparseoccvla,wang2026vggdrive}. In contrast, we focus on vision-only semantic occupancy and convert visual and geometric foundation features into sparse Gaussian cues.

% ===========================================================================
% ==== 3 Method ====
% ===========================================================================

\section{Method}

\subsection{Problem Statement} 

Given calibrated surround-view images $\mathcal{I}=\{I_c\}_{c=1}^{N}$ and camera parameters $\mathcal{P}=\{K_c,T_{c\rightarrow e}\}_{c=1}^{N}$, vision-only occupancy prediction estimates an ego-frame semantic field $\hat{\mathcal{Y}}$.
We describe a Gaussian occupancy representation by its locations $\mathcal{X}$ and attribute states $\mathcal{A}$, which are initialized and updated using feature inputs $\mathcal{K}$:
\begin{equation}
\mathcal{R}
=
\operatorname{Learn}\!\left(
\mathcal{X},\mathcal{A}\mid\mathcal{K}
\right),
\qquad
\hat{\mathcal{Y}}
=
D_{\theta}(\mathcal{R}).
\label{eq:occ_task_variables}
\end{equation}
Sparse Gaussian methods instantiate $\mathcal{R}$ with $M$ primitives and obtain update features by sampling camera features at projected 3D reference points:
\begin{equation}
\begin{aligned}
\mathcal{Q}
&=
\{Q_q\}_{q=1}^{M},\\
Q_q
&=
(\mu_q,\Sigma_q,\alpha_q,s_q,f_q),\\
k_q
&=
\left\{
F_c\!\left(\pi_c(x_{qn}^{\mathrm{ref}})\right)
\right\}_{c=1,j=1}^{N,J}.
\end{aligned}
\label{eq:gaussian_task_variables}
\end{equation}
Here, $x_{qn}^{\mathrm{ref}}$ denotes the $n$-th 3D reference point associated with primitive $q$. The centers $\mu_q$ provide compact 3D locations, while $\Sigma_q$, $\alpha_q$, $s_q$, and $f_q$ encode shape, occupancy strength, semantics, and latent content. However, generic query initialization provides limited scene-specific information, while projected sampling does not explicitly encode calibrated ray and cross-view relations for volumetric reasoning. Thus, sparse primitives alone are insufficient for effective Gaussian occupancy modeling.

Therefore, we incorporate visual and geometric cues into sparse Gaussian initialization and refinement. The desired Gaussian representation should obtain a geometry-guided center set $\mathcal{C}$, a visual semantic attribute set $\mathcal{V}$, and pose-aware refinement features $\mathcal{F}$. Visual and geometric foundation models provide geometric features $\mathcal{Z}_{\mathrm{geo}}$ and visual semantic features $\mathcal{Z}_{\mathrm{vis}}$. We formulate this process as:
\begin{equation}
\begin{aligned}
\left(
\mathcal{Z}_{\mathrm{geo}},
\mathcal{Z}_{\mathrm{vis}}
\right)
&=
\Phi_{\mathrm{VG}}(\mathcal{I}),\\
\left(
\mathcal{C},
\mathcal{V},
\mathcal{F}
\right)
&=
\Psi\!\left(
\mathcal{Z}_{\mathrm{geo}},
\mathcal{Z}_{\mathrm{vis}},
\mathcal{P}
\right),\\
\mathcal{G}
&=
\operatorname{Learn}\!\left(
\mathcal{C},
\mathcal{V}
\mid
\mathcal{F}
\right).
\end{aligned}
\label{eq:visual_geometric_representation}
\end{equation}
We term the resulting Gaussian set $\mathcal{G}$ \emph{Visual-Geometric Gaussians}. This formulation summarizes how visual and geometric cues support Gaussian initialization and refinement.

% ===========================================================================

\subsection{Overall Design}

Figure~\ref{fig:method} illustrates the VGOcc pipeline. Given surround-view images $\mathcal{I}$, frozen VGGT extracts multi-level DINO, frame, and global tokens as visual and geometric features, while camera parameters $\mathcal{P}$ provide calibrated ray and pose information. Frozen geometric features pass through a trainable DPT head to produce ray depth hypotheses. Non-empty candidates are lifted into 3D and selected by fast voxel-thinned sampling to form spatially balanced centers $\mathcal{C}$. Learned fallback queries cover weakly observed regions, while corresponding DINO features initialize primitive attributes $\mathcal{V}$. Together, $\mathcal{C}$ and $\mathcal{V}$ form the birth Gaussian queries $\mathcal{Q}$. At refinement stage $j$, foundation tokens are conditioned on camera embeddings and calibrated patch rays before fusion. Neighboring-view features are then aggregated at projected 3D locations to produce $F^{j}$. Four stages yield $\mathcal{F}=\{F^{j}\}_{j=1}^{4}$, which guides Gaussian Occupancy Decoder to refine $\mathcal{Q}$ into the final set $\mathcal{G}$, termed \emph{Visual-Geometric Gaussians}. Finally, $\mathcal{G}$ is rendered into the semantic occupancy $\hat{\mathcal{Y}}$.

% ===========================================================================
\subsection{Visual-Geometric Gaussian Birth}

We design Visual-Geometric Gaussian Birth to construct the center set $\mathcal{C}$ and attribute set $\mathcal{V}$ defined in Eq.~\ref{eq:visual_geometric_representation}. Generic learned queries provide limited scene-specific placement, while confidence ranking tends to concentrate primitives near dominant visible surfaces. We use ray depth hypotheses to generate scene-specific candidates, fast voxel-thinned sampling to maintain spatial coverage, and visual features to initialize primitive attributes.

\paragraph{Ray-depth hypothesis construction.}
For patch $i$ in camera $c$, let $(o_c,r_{ci})$ denote its calibrated ego-frame ray. A lightweight DPT head predicts a depth posterior with an empty category $p_{\mathrm{ray}}(b\mid c,i)$ from frozen geometric features. The empty category suppresses rays without occupied structures. Each non-empty depth bin produces:
\begin{equation}
\begin{aligned}
  x_{cib} &= o_c+d_b r_{ci},
  & w_{cib} &= p_{\mathrm{ray}}(b\mid c,i),\\
  \xi_{ci} &= \phi_{\mathrm{DINO}}(c,i),
  & b &\neq\emptyset .
\end{aligned}
\label{eq:birth_candidate}
\end{equation}
Here $x_{cib}$ is a 3D center candidate, $w_{cib}$ is its confidence, and $\xi_{ci}$ is the DINO token from the same ray. Pairing each candidate with its visual token preserves the correspondence between geometric placement and visual semantics without affecting its location.

\paragraph{Fast voxel-thinned sampling.}
Global confidence ranking favors visible and near-camera regions. We instead partition the candidates into coarse voxels:
\begin{equation}
  \nu(x_i)
  =
  \left\lfloor
  \frac{x_i-p_{\min}}{s_v}
  \right\rfloor.
  \label{eq:birth_voxel_sampling}
\end{equation}
Non-empty voxels are sampled uniformly, while candidates within each selected voxel are sampled according to their confidence $w_i$. This balances spatial coverage and geometric reliability, producing $M_g$ geometry-guided queries. Another $M_p$ learned fallback queries cover weakly observed regions, with $M=M_g+M_p$.

\paragraph{Pose-aware attribute initialization.}
After geometry determines the centers, visual features initialize the non-spatial primitive attributes. For a geometry-guided query, $\Omega_q$ contains its source patch. For a fallback query, it contains valid projections into visible cameras. The corresponding features are aggregated as:
\begin{equation}
\begin{aligned}
  \hat{\xi}_q
  &=
  \operatorname{Norm}_{\mathcal{Q}}\!\left(
  \operatorname{Avg}_{\Omega_q}
  \left[
  \operatorname{PoseGate}(\xi_{ci},\rho_{ci})
  \right]
  \right),\\
  a_q
  &=
  a_{\mathrm{base}}
  +
  \boldsymbol{\eta}\odot
  \operatorname{MLP}_{a}(\hat{\xi}_q).
\end{aligned}
\label{eq:birth_attributes}
\end{equation}
Here $a_q=(\alpha_q,s_q,f_q)$ contains the opacity, semantic, and latent feature states. $\operatorname{MLP}_{a}$ denotes the attribute prediction heads, and $\boldsymbol{\eta}$ contains their residual scales. The covariance $\Sigma_q$ is initialized from learned bases. Residual prediction retains the base initialization while introducing scene-specific visual semantics. The resulting Gaussian queries are:
\begin{equation}
\begin{aligned}
  Q_q
  &=
  (\mu_q,\Sigma_q,\alpha_q,s_q,f_q),\\
  \mathcal{C}
  &=
  \{\mu_q\}_{q=1}^{M},\\
  \mathcal{V}
  &=
  \{(\Sigma_q,\alpha_q,s_q,f_q)\}_{q=1}^{M},\\
  \mathcal{Q}
  &=
  \{Q_q\}_{q=1}^{M}
  =
  \mathcal{Q}_g\cup\mathcal{Q}_p.
\end{aligned}
\label{eq:birth_output}
\end{equation}
Visual-Geometric Gaussian Birth therefore uses geometry for Gaussian placement, balanced allocation for spatial coverage, and visual semantics for attribute initialization.

\definecolor{vgTabBarrier}{RGB}{255,115,45}
\definecolor{vgTabBicycle}{RGB}{245,185,205}
\definecolor{vgTabBus}{RGB}{245,245,0}
\definecolor{vgTabCar}{RGB}{20,150,230}
\definecolor{vgTabConstVeh}{RGB}{0,210,210}
\definecolor{vgTabMotorcycle}{RGB}{175,175,0}
\definecolor{vgTabPedestrian}{RGB}{245,0,0}
\definecolor{vgTabTrafficCone}{RGB}{255,235,145}
\definecolor{vgTabTrailer}{RGB}{120,55,0}
\definecolor{vgTabTruck}{RGB}{165,45,220}
\definecolor{vgTabDriveSurf}{RGB}{245,0,220}
\definecolor{vgTabOtherFlat}{RGB}{150,130,115}
\definecolor{vgTabSidewalk}{RGB}{80,0,80}
\definecolor{vgTabTerrain}{RGB}{110,225,80}
\definecolor{vgTabManmade}{RGB}{215,215,215}
\definecolor{vgTabVegetation}{RGB}{0,160,0}

\providecommand{\vgtabclsbox}[1]{\textcolor{#1}{\rule{2mm}{2mm}}}
\providecommand{\vgtabheadlower}[1]{%
  \vphantom{#1}\raisebox{-2mm}[0pt][0pt]{#1}%
}
\providecommand{\vgtabrotcls}[2]{%
  \vgtabheadlower{%
    \begin{tabular}[b]{@{}c@{}}
      \rotatebox[origin=c]{90}{\scriptsize #1}\\[-0.4mm]
      \vgtabclsbox{#2}
    \end{tabular}%
  }%
}

\begin{table*}[!t]
\centering
\scriptsize
\setlength{\tabcolsep}{2.8pt}
\renewcommand{\arraystretch}{1.12}
\resizebox{\textwidth}{!}{%
\begin{tabular}{@{}l|c|cc|cccccccccccccccc@{}}
\toprule
\vgtabheadlower{\textbf{Method}}
&
\vgtabheadlower{\textbf{Venue}}
&
\begin{tabular}[c]{@{}c@{}}\textbf{SC}\\\textbf{IoU}\end{tabular}
&
\begin{tabular}[c]{@{}c@{}}\textbf{SSC}\\\textbf{mIoU}\end{tabular}
&
\vgtabrotcls{barrier}{vgTabBarrier}
&
\vgtabrotcls{bicycle}{vgTabBicycle}
&
\vgtabrotcls{bus}{vgTabBus}
&
\vgtabrotcls{car}{vgTabCar}
&
\vgtabrotcls{const. veh.}{vgTabConstVeh}
&
\vgtabrotcls{motorcycle}{vgTabMotorcycle}
&
\vgtabrotcls{pedestrian}{vgTabPedestrian}
&
\vgtabrotcls{traffic cone}{vgTabTrafficCone}
&
\vgtabrotcls{trailer}{vgTabTrailer}
&
\vgtabrotcls{truck}{vgTabTruck}
&
\vgtabrotcls{drive. suf.}{vgTabDriveSurf}
&
\vgtabrotcls{other flat}{vgTabOtherFlat}
&
\vgtabrotcls{sidewalk}{vgTabSidewalk}
&
\vgtabrotcls{terrain}{vgTabTerrain}
&
\vgtabrotcls{manmade}{vgTabManmade}
&
\vgtabrotcls{vegetation}{vgTabVegetation}
\\
\midrule
MonoScene
& CVPR'22
& 23.96 & 7.31
& 4.03 & 0.35 & 8.00 & 8.04 & 2.90 & 0.28 & 1.16 & 0.67
& 4.01 & 4.35 & 27.72 & 5.20 & 15.13 & 11.29 & 9.03 & 14.86
\\
Atlas
& ECCV'20
& 28.66 & 15.00
& 10.64 & 5.68 & 19.66 & 24.94 & 8.90 & 8.84 & 6.47 & 3.28
& 10.42 & 16.21 & 34.86 & 15.46 & 21.89 & 20.95 & 11.21 & 20.54
\\
BEVFormer
& ECCV'22
& 30.50 & 16.75
& 14.22 & 6.58 & 23.46 & 28.28 & 8.66 & 10.77 & 6.64 & 4.05
& 11.20 & 17.78 & 37.28 & 18.00 & 22.88 & 22.17 & 13.80 & \underline{22.21}
\\
TPVFormer
& CVPR'23
& 11.51 & 11.66
& 16.14 & 7.17 & 22.63 & 17.13 & 8.83 & 11.39 & 10.46 & 8.23
& 9.43 & 17.02 & 8.07 & 13.64 & 13.85 & 10.34 & 4.90 & 7.37
\\
TPVFormer$^\dagger$
& CVPR'23
& 30.86 & 17.10
& 15.96 & 5.31 & 23.86 & 27.32 & 9.79 & 8.74 & 7.09 & 5.20
& 10.97 & 19.22 & 38.87 & 21.25 & 24.26 & 23.15 & 11.73 & 20.81
\\
OccFormer
& ICCV'23
& 31.39 & 19.03
& 18.65 & 10.41 & 23.92 & \underline{30.29} & 10.31 & 14.19 & 13.59 & 10.13
& 12.49 & 20.77 & 38.78 & 19.79 & 24.19 & 22.21 & 13.48 & 21.35
\\
SurroundOcc
& ICCV'23
& \underline{31.49} & \underline{20.30}
& \textbf{20.59} & 11.68 & 28.06 & \textbf{30.86} & 10.70 & 15.14 & \textbf{14.09} & \textbf{12.06}
& \underline{14.38} & \underline{22.26} & 37.29 & \underline{23.70} & 24.49 & 22.77 & \underline{14.89} & 21.86
\\
GaussianFormer
& ECCV'24
& 29.83 & 19.10
& 19.52 & 11.26 & 26.11 & 29.78 & 10.47 & 13.83 & 12.58 & 8.67
& 12.74 & 21.57 & \underline{39.63} & 23.28 & 24.46 & 22.99 & 9.59 & 19.12
\\
GaussianFormer-2
& CVPR'25
& 30.56 & 20.02
& 20.15 & \textbf{12.99} & 27.61 & 30.23 & 11.19 & \underline{15.31} & 12.64 & 9.63
& 13.31 & \underline{22.26} & 39.68 & 23.47 & \underline{25.62} & \underline{23.20} & 12.25 & 20.73
\\
\midrule
\textbf{VGOcc}
& \textbf{Ours}
& \textbf{34.07} & \textbf{21.75}
& \underline{20.43} & \underline{12.48} & \textbf{29.27} & 30.00 & \textbf{13.24} & \textbf{16.10} & \underline{13.97} & \underline{10.50}
& \textbf{15.09} & \textbf{22.50} & \textbf{43.09} & \textbf{25.93} & \textbf{28.17} & \textbf{26.21} & \textbf{16.22} & \textbf{24.83}
\\
\bottomrule
\end{tabular}
}
\caption{
3D semantic occupancy prediction results on nuScenes. $^\dagger$ means supervised by dense occupancy annotations as opposed to original LiDAR segmentation labels. The best and second-best results are represented by \textbf{bold} and \underline{underline} respectively.
}
\label{tab:main}
\end{table*}

\subsection{Pose-Aware Feature Learning}

Visual-Geometric Gaussian Birth constructs the query set $\mathcal{Q}$ once, whereas the four decoder stages operate at different feature levels. Raw foundation tokens lack patch-level ray identity and calibrated cross-view consistency. We therefore apply Pose-Aware Feature Learning at each stage $j$ to produce stage-specific feature maps $F^j$.

\paragraph{Patch-local pose conditioning.}
A camera-level embedding assigns the same pose information to all patches in one view, although each patch corresponds to a distinct ray. For patch $i$ in camera $c$, we encode:
\begin{equation}
\begin{aligned}
  \rho_{ci}
  &=
  [r_{ci},\,o_c\times r_{ci},\,o_c,\,\bar{u}_i],\\
  t_{ci}
  &=
  E_{\mathrm{ray}}\!\left(\Gamma(\rho_{ci})\right),
\end{aligned}
\label{eq:pafl_pose_ray}
\end{equation}
where $\bar{u}_i$ is the normalized patch coordinate and $\Gamma$ denotes Fourier encoding. The resulting $t_{ci}$ distinguishes each patch by its calibrated origin, direction, and image location.

\paragraph{Stage-wise tri-stream fusion.}
At stage $j$, foundation level $\ell_j$ provides DINO, frame, and global tokens $z_{ci}^{\ell_j,\tau}$, where $\tau\in\{d,f,g\}$. These streams provide semantic, local geometric, and global contextual cues. Since their contributions vary across refinement stages, each stream is projected and conditioned on the local pose before fusion:
\begin{equation}
\begin{aligned}
  \tilde{z}_{ci}^{j,\tau}
  &=
  \operatorname{PoseFiLM}_{j,\tau}
  \left(
  P_{\tau}z_{ci}^{\ell_j,\tau},
  t_{ci}
  \right),\\
  h_{ci}^{j}
  &=
  \operatorname{Fuse}_{j}
  \left(
  \{\tilde{z}_{ci}^{j,\tau}\}_{\tau},
  t_{ci}
  \right).
\end{aligned}
\label{eq:pafl_stream_fusion}
\end{equation}
Here $\operatorname{Fuse}_{j}$ is a pose-conditioned softmax gate over the three streams. It preserves their complementary information while adapting their contributions to each stage.

\paragraph{Pose-guided cross-view aggregation.}
Patch-local pose identifies the source ray but does not resolve its depth ambiguity. We therefore sample metric points $X_{cik}=o_c+\delta_k r_{ci}$ and project them into neighboring cameras. Let $y_{c'ik}^{j}$ denote a sampled target feature and $\gamma_{c'k}=\max(r_{ci}^{\top}r_{c'ik},0)$ its ray compatibility. Cross-view aggregation is defined as:
\begin{equation}
\begin{aligned}
  e_{c'k}^{j}
  &=
  \frac{
  \langle W_q^{j}h_{ci}^{j},
  W_k^{j}y_{c'ik}^{j}\rangle
  }{\sqrt{d_h}}
  +
  \lambda_a\log(\gamma_{c'k}+\epsilon),\\
  \Delta h_{ci}^{j}
  &=
  \sum_{c',k}
  a_{c'k}^{j}W_v^{j}y_{c'ik}^{j},
  \qquad
  a^{j}
  =
  \operatorname{softmax}(e^{j}).
\end{aligned}
\label{eq:pafl_cross_view}
\end{equation}
Invalid projections are masked. Combining feature similarity with ray compatibility suppresses geometrically inconsistent projections and introduces depth-aware cross-view information.

The fused source and cross-view features are converted into the maps used by each decoder stage:
\begin{equation}
\begin{aligned}
  F_c^{j}
  &=
  \operatorname{Map}_{j}
  \left(
  \{\operatorname{LN}
  (h_{ci}^{j}+W_o^{j}\Delta h_{ci}^{j})\}_{i}
  \right),\\
  \mathcal{F}
  &=
  \left\{
  \{F_c^{j}\}_{c=1}^{N}
  \right\}_{j=1}^{4}.
\end{aligned}
\label{eq:pafl_output}
\end{equation}
Thus, $\mathcal{F}$ provides stage-specific pose-conditioned and cross-view features for refining $\mathcal{Q}$ into the final Gaussian set $\mathcal{G}$.

% ===========================================================================
\subsection{Gaussian Occupancy Decoder}

The decoder takes the birth Gaussian queries $\mathcal{Q}$ and the pose-aware features $\mathcal{F}=\{F^{j}\}_{j=1}^{4}$. At stage $j$, the current Gaussian centers are projected into the camera views to sample $F^{j}$, which refines their centers, covariance, opacity, semantics, and latent features. After four stages, the decoder produces the final Gaussian set $\mathcal{G}$ and renders it into semantic occupancy:
\begin{equation}
\begin{aligned}
  \mathcal{G}
  &=
  \mathcal{D}_{\mathrm{SG}}(\mathcal{Q},\mathcal{F})
  =
  \left\{
  (\bar{\mu}_q,\bar{\Sigma}_q,
  \bar{\alpha}_q,\bar{s}_q,\bar{f}_q)
  \right\}_{q=1}^{M},\\
  \hat{\mathcal{Y}}
  &=
  \operatorname{GS2Occ}(\mathcal{G}).
\end{aligned}
\label{eq:gaussian_occ_decoder}
\end{equation}
We retain the standard sparse Gaussian refinement and rendering process, while modifying only the initial queries $\mathcal{Q}$ and stage-specific features $\mathcal{F}$. This keeps the decoder unchanged and attributes the resulting gains to Gaussian birth and pose-aware feature learning.

% ===========================================================================
\subsection{Training Details}

VGOcc is trained with the visual geometry foundation model frozen. The trainable components include the DPT ray depth head, visual attribute heads, Pose-Aware Feature Learning blocks, Gaussian Occupancy Decoder, and GS2Occ head. Given the predicted semantic occupancy field $\hat{\mathcal{Y}}$ and ground truth $\mathcal{Y}$, the overall objective is:
\begin{equation}
  \mathcal{L}
  =
  \mathcal{L}_{\mathrm{occ}}
  +\lambda_{\mathrm{pix}}\mathcal{L}_{\mathrm{pix}}
  +\lambda_{\mathrm{aux}}\mathcal{L}_{\mathrm{aux}}.
  \label{eq:training_objective}
\end{equation}
The occupancy loss is defined as:
\begin{equation}
  \mathcal{L}_{\mathrm{occ}}
  =
  10\,\operatorname{CE}(\hat{\mathcal{Y}},\mathcal{Y})
  +
  \operatorname{Lovasz}(\hat{\mathcal{Y}},\mathcal{Y}).
\end{equation}
$\mathcal{L}_{\mathrm{pix}}$ supervises the ray depth posterior $p_{\mathrm{ray}}$ using the ray depth target $\mathcal{Y}_{\mathrm{ray}}$, while $\mathcal{L}_{\mathrm{aux}}$ denotes the auxiliary occupancy losses applied to intermediate decoder stages. We train VGOcc for 20 epochs using Adam with a learning rate of $4\times10^{-4}$ on four NVIDIA H20 GPUs.

% ===========================================================================
% ==== 4 Experiments ====
% ===========================================================================

% ==== Figure ====
\begin{figure*}[!t]
  \centering
  \includegraphics[width=1\textwidth]{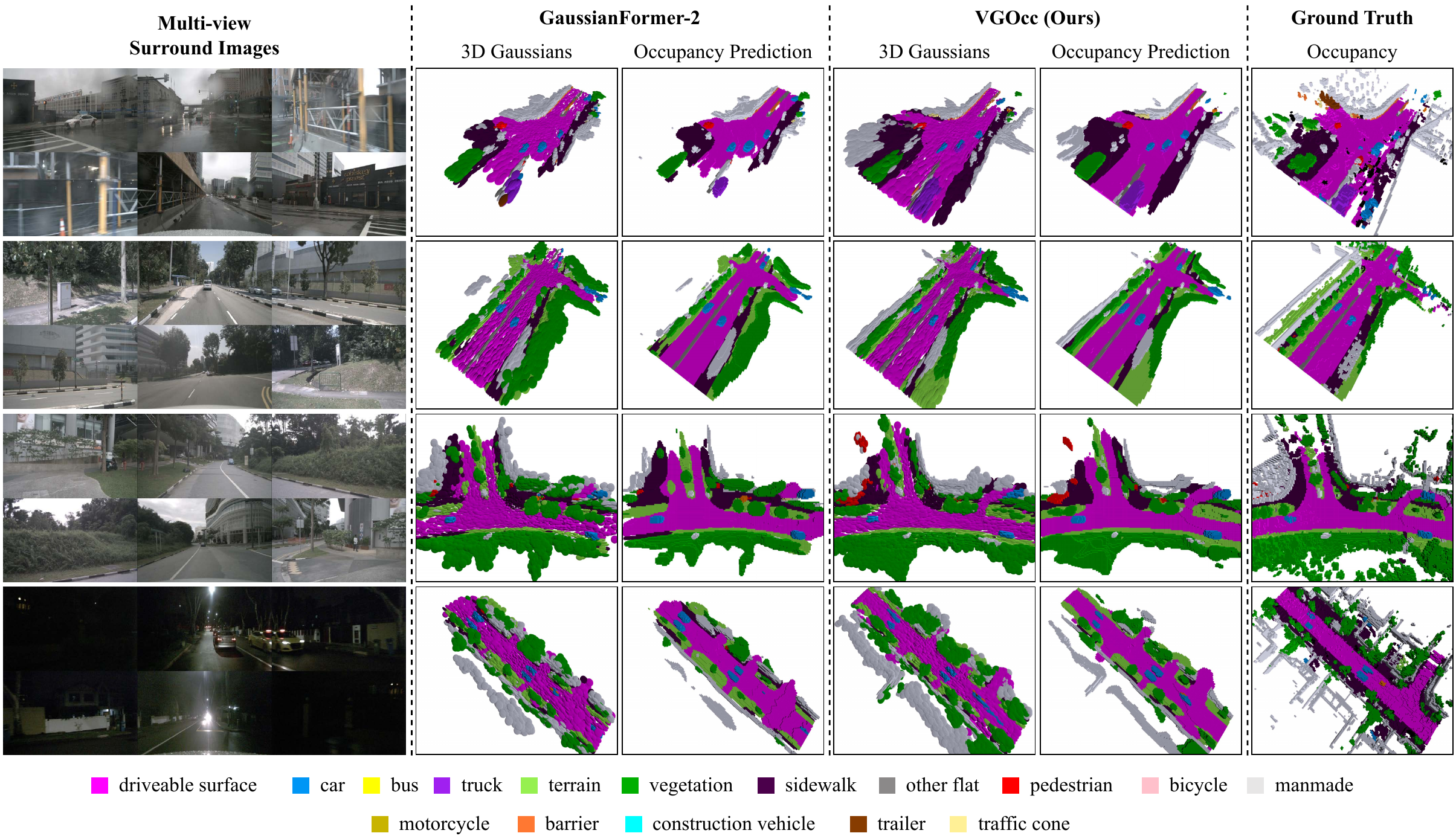}
  \caption{Qualitative results of our method and GaussianFormer-2 on SurroundOcc dataset.}
  \label{fig:exp1}
\end{figure*}

\section{Experiments}

\subsection{Experimental Setup}

\paragraph{Dataset.}
We conduct all experiments on nuScenes \citep{caesar2020nuscenes}, which contains 1,000 driving scenes recorded in Boston and Singapore. The official split contains 700 training, 150 validation, and 150 testing scenes, with synchronized images from six surround-view cameras. Following SurroundOcc \citep{wei2023surroundocc}, we use its 3D semantic occupancy labels for training and evaluation. The ego-centered evaluation region spans $100\,\mathrm{m}\times100\,\mathrm{m}\times8\,\mathrm{m}$, with $X,Y\in[-50\,\mathrm{m},50\,\mathrm{m}]$ and $Z\in[-5\,\mathrm{m},3\,\mathrm{m}]$. Discretization at $0.5\,\mathrm{m}$ produces a $200\times200\times16$ voxel grid with 16 semantic classes, one empty class, and one unknown class.

\paragraph{Metrics.} 
Following standard semantic occupancy evaluation \citep{cao2022monoscene}, we report semantic scene completion mIoU and scene completion IoU. The former averages class-wise IoU over occupied semantic classes $\mathcal{C}_{\mathrm{sem}}$:
\begin{equation}
  \mathrm{mIoU}
  =
  \frac{1}{|\mathcal{C}_{\mathrm{sem}}|}
  \sum_{i\in\mathcal{C}_{\mathrm{sem}}}
  \frac{\mathrm{TP}_i}
  {\mathrm{TP}_i+\mathrm{FP}_i+\mathrm{FN}_i}.
  \label{eq:miou_metric}
\end{equation}
For scene completion, all non-empty classes are merged into one occupied class $o$:
\begin{equation}
  \mathrm{IoU}
  =
  \frac{\mathrm{TP}_o}
  {\mathrm{TP}_o+\mathrm{FP}_o+\mathrm{FN}_o}.
  \label{eq:iou_metric}
\end{equation}
Here $\mathrm{TP}$, $\mathrm{FP}$, and $\mathrm{FN}$ denote true positives, false positives, and false negatives.

\paragraph{Compared methods.}
We compare VGOcc with representative vision-only occupancy methods under the same nuScenes evaluation protocol. The structured representation baselines include MonoScene \citep{cao2022monoscene}, Atlas \citep{murez2020atlas}, BEVFormer \citep{li2022bevformer}, TPVFormer \citep{huang2023tpvformer}, VoxFormer \citep{li2023voxformer}, OccFormer \citep{zhang2023occformer}, SurroundOcc \citep{wei2023surroundocc}, COTR \citep{ma2024cotr}, and SparseOcc \citep{liu2024sparseocc}. We further compare with the sparse Gaussian methods GaussianFormer \citep{huang2024gaussianformer} and GaussianFormer-2 \citep{huang2025gaussianformer2}. All results are taken from official vison-only evaluations when available.

\paragraph{Implementation details.}
VGOcc uses frozen VGGT to extract multi-level DINO, frame, and global tokens \citep{wang2025vggt}. DINO tokens provide visual-semantic features for Gaussian attributes and stage-wise feature learning \citep{oquab2024dinov2}. The foundation feature grid is $21\times37$. We fix the Gaussian budget at $25.6$k primitives, including $19.2$k geometry-guided queries and $6.4$k learned fallback queries. The trainable components are the DPT ray-posterior head, visual attribute heads, four Pose-Aware Feature Learning blocks, sparse Gaussian decoder, and Gaussian-to-voxel head. All ablation studies retain the same Gaussian budget and decoder unless otherwise stated.

\begin{figure*}[!t]
  \centering
  \includegraphics[width=1\textwidth]{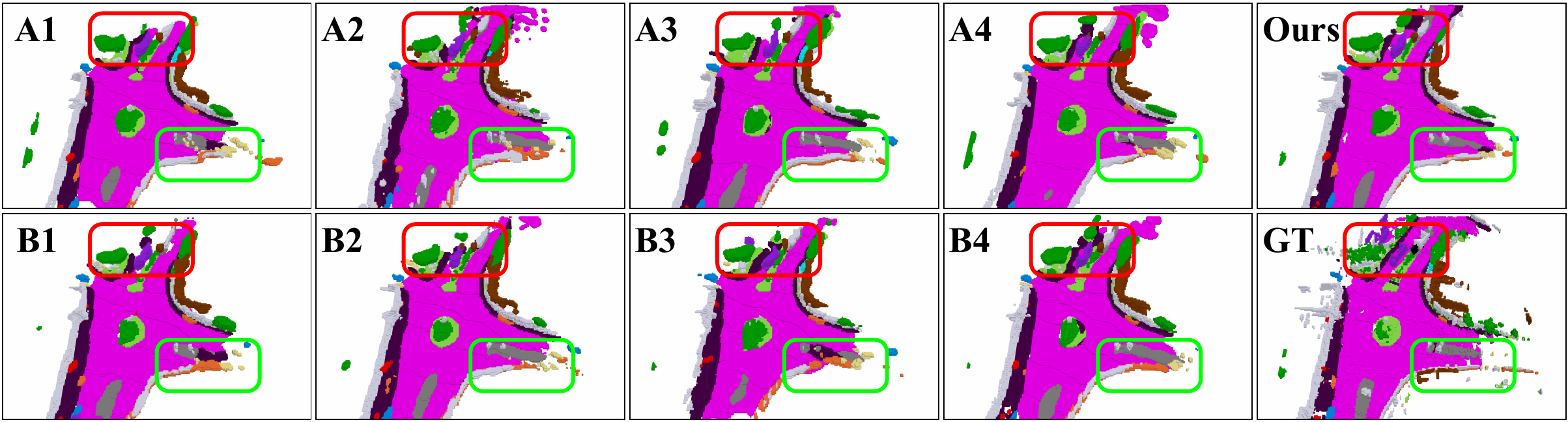}
  \caption{Qualitative results of ablation studies. Red and Green rectangular boxes highlight the differences.}
  \label{fig:exp2}
\end{figure*}

% ===========================================================================
\subsection{Quantitative Results}

As shown in Table~\ref{tab:main}, VGOcc achieves 34.07 SC IoU and 21.75 SSC mIoU, surpassing non-Gaussian baseline SurroundOcc by 2.58 and 1.45 points. It ranks first among non-Gaussian methods in 12 of 16 classes. Clear gains on large-scale surfaces and foreground objects show that VGOcc captures scene geometry and semantic content better than BEV, tri-plane, and voxel representations. This indicates that the visual-geometric representation preserves global layout while retaining object-level semantics. Lower scores on barrier, car, pedestrian, and traffic cone suggest that compact or thin classes remain sensitive to limited image support and sparse sampling. Overall, VGOcc provides a more balanced representation of spatial structure and semantic detail.

Compared with GaussianFormer-2, VGOcc improves SC IoU by 3.51 and SSC mIoU by 1.73 points, with higher scores in 14 of 16 classes. The broad gains across large-scale surfaces and foreground objects demonstrate the effectiveness of the overall visual-geometric design in improving both structural recovery and semantic modeling. The consistent improvements across categories further show that these benefits are not limited to either background layout or discrete objects. Since both methods use sparse semantic Gaussians and the same decoder, the results confirm that scene-specific visual and geometric cues produce a stronger Gaussian representation than sparsity alone.

% ==== Table ====
\begin{table}[t]
\centering
\caption{Quantitative results of ablation studies. All variants use the same Gaussian budget and decoder.}
\label{tab:abs1}
\setlength{\tabcolsep}{5pt}
\renewcommand{\arraystretch}{1.08}
\resizebox{\columnwidth}{!}{
\begin{tabular}{llcc}
\toprule
Variant & Modification & IoU $\uparrow$ & mIoU $\uparrow$ \\
\midrule
\multicolumn{4}{l}{\textit{Visual-Geometric Gaussian Birth}} \\
A1 & w/o visual residual initialization & 33.64 & 21.09 \\
A2 & GaussianFormer-2 query initialization & 32.42 & 20.24 \\
A3 & w/o pose-ray attribute gate & 33.78 & 21.46 \\
A4 & FPS instead of Fast Voxel-thinned Sampling & 33.74 & 21.23 \\
\midrule
\multicolumn{4}{l}{\textit{Pose-Aware Feature Learning}} \\
B1 & w/o patch-local pose conditioning & 32.75 & 21.14 \\
B2 & w/o DINO stream & 33.27 & 21.06 \\
B3 & independent adapters at all levels & 31.87 & 19.96 \\
B4 & w/o pose-guided cross-view aggregation & 33.67 & 21.37 \\
\midrule
\multicolumn{2}{l}{\textbf{Full VGOcc}} & \textbf{34.07} & \textbf{21.75} \\
\bottomrule
\end{tabular}}
\end{table}

\subsection{Qualitative Results}
Figure~\ref{fig:exp1} compares VGOcc with GaussianFormer-2 under rainy, sunny, cloudy, and nighttime conditions. All Gaussian and occupancy results are rendered from the same viewpoint. In the 2nd row sunny and 3rd row cloudy scenes, VGOcc preserves clearer road boundaries, sidewalks, and vegetation, producing more complete occupancy predictions. In the 1st row rainy and 4th row nighttime scenes, GaussianFormer-2 suffers severe structural failures and scattered errors, while VGOcc retains coherent 3D geometry and scene layout. These results show that \emph{Visual-Geometric Gaussians} provide a stronger representation for semantic occupancy across diverse conditions.

% ===========================================================================

\subsection{Ablation Studies}

We conduct ablation studies to verify that VGOcc effectively learns visual and geometric cues for constructing \emph{Visual-Geometric Gaussians}. All variants are evaluated under the same experimental setting for fair comparison.

\paragraph{Visual-Geometric Gaussian Birth.}
Table~\ref{tab:abs1} evaluates the proposed Gaussian birth strategy. Removing visual residual initialization (\textbf{A1}) decreases IoU and mIoU by 0.43 and 0.66, confirming that same-ray visual features provide scene-specific semantic priors. Replacing VGBirth with GaussianFormer-2 initialization (\textbf{A2}) causes the largest drops of 1.65 and 1.51, demonstrating the joint contribution of ray-depth centers, visual attributes, and balanced allocation. Removing the pose-ray attribute gate (\textbf{A3}) reduces both metrics by 0.29, showing that calibrated camera and ray cues improve attribute initialization. Replacing fast voxel-thinned sampling with FPS (\textbf{A4}) decreases IoU and mIoU by 0.33 and 0.52, validating our sampling strategy for balanced center allocation. 
As shown in Figure~\ref{fig:exp2}, the full design preserves clearer road boundaries and structures than variants \textbf{A1-A4}. Overall, VGBirth provides effective geometric and semantic initialization for sparse Gaussians.

\paragraph{Pose-Aware Feature Learning.}
The second group evaluates feature learning during Gaussian refinement. Removing patch-local pose conditioning (\textbf{B1}) decreases IoU and mIoU by 1.32 and 0.61, confirming that patch-specific ray geometry is essential for feature adaptation. Removing the DINO stream (\textbf{B2}) causes drops of 0.80 and 0.69, showing that visual semantics complement geometric and global cues. Using independent adapters at all levels (\textbf{B3}) produces the largest degradation of 2.20 and 1.79, demonstrating that shared adaptation improves feature consistency across levels. Removing pose-guided cross-view aggregation (\textbf{B4}) reduces IoU and mIoU by 0.40 and 0.38, validating calibrated neighboring-view features for resolving ray-depth ambiguity. 
Figure~\ref{fig:exp2} also shows that the full design yields cleaner semantic boundaries and fewer fragmented regions than variants \textbf{B1-B4}. Overall, PFLearn provides effective pose-aware and multi-view features for Gaussian refinement.

% ===========================================================================
% ==== 5 Conclusion ====
% ===========================================================================

\section{Conclusion}

We present VGOcc, a vision-centric 3D occupancy framework that learns \emph{Visual-Geometric Gaussians} from frozen visual and geometric foundation features. First, we form scene-specific centers through ray-depth hypotheses and fast voxel-thinned sampling, while same-ray visual features initialize primitive attributes. Then, we condition multi-level foundation tokens on calibrated pose-ray information and aggregate neighboring-view features across multiple stages for iterative refinement. Together, these designs improve scene structure recovery and semantic modeling, achieving state-of-the-art performance on nuScenes and robust predictions across diverse imaging conditions.

VGOcc still has several limitations, including its reliance on accurate cameras, the computational overhead of the frozen backbone, and the lack of temporal modeling. Future work will focus on addressing these challenges.

\section{Acknowledgments}
This work was supported by National Science and Technology Major Project (2024ZD01NL00101), Natural Science Foundation of China (62271013), Guangdong Provincial Key Laboratory of Ultra High Definition Immersive Media Technology (2024B1212010006), Guangdong Province Pearl River Talent Program (2021QN020708), Guangdong Basic and Applied Basic Research Foundation (2024A1515010155), Shenzhen Science and Technology Program (JCYJ20240813160202004, JCYJ20230807120808017, SYSPG20241211173440004), Shenzhen Fundamental Research Program (GXWD20201231165807007-20200806163656003), and financially supported for Outstanding Talents Training Fund in Shenzhen. We also thank Great Wall Motor Co., Ltd. for their support.
% ===========================================================================
% ==== References ====
% ===========================================================================
\bibliography{refs}
\clearpage
\end{document}